ORIGINAL RESEARCH PAPER

# An eco-driving approach for ride comfort improvement

Ó. Mata-Carballeira | I. del Campo | E. Asua

Department of Electricity and Electronics, Faculty of Science and Technology, University of the Basque Country, Spain

**Correspondence**
Ó. Mata-Carballeira, Department of Electricity and Electronics, Faculty of Science and Technology, University of the Basque Country UPV/EHU, 48940 Leioa, Spain.
Email: oscar.mata@ehu.eus

**Funding information**
University of the Basque Country UPV/EHU, Grant/Award Number: GIU18/122; European Commission, Grant/Award Number: TEC2016-77618-R; Spanish AEI, Grant/Award Number: TEC2016-77618-R; Basque Government, Grant/Award Number: KK-2019-00035-AUTOLIB (ELKARTEK)

**Abstract**
New challenges on transport systems are emerging due to the advances that the current paradigm is experiencing. The breakthrough of the autonomous car brings concerns about ride comfort, while the pollution concerns have arisen in recent years. In the model of automated automobiles, drivers are expected to become passengers, so, they will be more prone to suffer from ride discomfort or motion sickness. Conversely, the eco-driving implications should not be set aside because of the influence of pollution on climate and people's health. For that reason, a joint assessment of the aforementioned points would have a positive impact. Thus, this work presents a self-organised map-based solution to assess ride comfort features of individuals considering their driving style from the viewpoint of eco-driving. For this purpose, a previously acquired dataset from an instrumented car was used to classify drivers regarding the causes of their lack of ride comfort and eco-friendliness. Once drivers are classified regarding their driving style, natural-language-based recommendations are proposed to increase the engagement with the system. Hence, potential improvements of up to the 57.7% for ride comfort evaluation parameters, as well as up to the 47.1% in greenhouse-gasses emissions are expected to be reached.

## 1 | INTRODUCTION

The paradigm of transportation is experiencing technological advances, causing new challenges to emerge [1]. On the one hand, the breakthrough of the autonomous driving scenario aims to relieve motorists from the tasks and risks of driving a car, with the subsequent improvement of safety and perceived comfort [2, 3]. Thus, while the former is expected to be clearly enhanced since most of the road accident-associated injuries and fatalities depend on the human factor [4], mostly due to recklessness and distractions [5], the latter requires further attention. This change of the driving scenario will turn drivers into mere passengers, and hence, new challenges to their wellness will appear [6], being passenger comfort one of them. On the other hand, despite pollution having been an issue since the popularisation of the automobile, it has become an even more concerning point in recent years due to the global warming, and because of the harmful effects on people's health and lives [7]. In consequence, eco-driving emerges as a set of rules and parameters to be followed with the aim of improving the energy efficiency while reducing the greenhouse gasses (GHG) emissions [8]. Hence, any autonomous car related development should take into account these rules to minimise both the impact of pollution and passenger's discomfort.

There exist many factors that contribute to the global ride comfort. These aspects are derived from both the environment within the vehicle (e.g. smell, temperature, humidity etc.), as well as from the characteristics of the passenger (e.g. gender, age, health conditions etc.) along with his/her behaviour (keeping or not his/her gaze on the road). Besides, vibrations affect the overall comfort perception. Several pieces of research link perceived distress, motion sickness and the frequency components of the vibrations with the resonance frequencies of the organs of the human body [9, 10]. Although some vibrations are caused by the constructive characteristics of either the roadway or the vehicle itself including the handling of the automobile, those derived from the driving behaviour and handling skills of the driver have a noticeable influence in compromising ride comfort.

The level of perceived discomfort is associated with the frequency of the vibrations the human body is exposed to,

and is directly proportional to their magnitude. It is worth to remark that increasing the time exposure of the individual to a given vibration source increases the discomfort sensation. With these assumptions in mind, it is known that low-frequency vibrations, with frequencies close to 1 Hz, propagate through the human body, impairing the well-being of the individual gradually until a no-return point. On the other hand, higher-frequency vibrations, which are significantly attenuated by the subject's body, have no significant contribution on increasing the sensation of malaise, but, on the other hand, they are related to a general feeling of disturbance. It is also interesting to note that monotone continuous low frequency vibrations increase fatigue, while transient vibrations produce stress [11].

Motion sickness is the most severe vibration-originated comfort condition, causing from mild effects such as cold sweating and dizziness to nausea and vomit, which strongly impair passengers [12, 13]. This condition, highly related to low-frequency vibrations, happens because the brain perceives a mismatch between the expected movements and the real accelerations detected by the vestibular apparatus of the inner ear [14]. For that reason, the vast majority of drivers do not get motion-sick, since they receive a more complete range of stimuli that helps the brain to have a more accurate insight into the actual dynamics of the car [15]. Furthermore, motion sickness is very frequent among passengers due to the fact of not perceiving as many stimuli as drivers do, consequently making the sensory conflict between the sight and the hearing senses more noticeable. It is worth noting that these effects are worsened as passengers get involved into secondary tasks, such as surfing the internet, reading or working on a PC [16].

At the same time, the concern caused by pollution is motivated by two main reasons: the GHG being the main actor in the global warning, and the harm that other associated toxic gasses cause. Regarding the latter point, it has been observed that people who suffer from pathologies of the breathing system tend to experience the most noticeable worsening of their health condition when exposed to high levels of pollutants [17]. Because of the both aforementioned reasons, several regulations intended to reducing the emission levels of private transportation have been passed by authorities, becoming their fulfilment one of the main objectives for the automotive industry [18–20].

Consequently, several works have put the spotlight on driving style (DS) as a major determinant on several handling features [21]. Thus, it is known that drivers have different DS, experiences and emotions due to unique driving characteristics, showing their own driving behaviours and habits [21]. Several pieces of research have boarded the assessment of DS to identify a variety of driving patterns, such as distracted driving, drowsiness or driving fatigue. Besides, works have been carried out on aggressive driving behaviour, which leads to several drawbacks for both the driver and the occupants, such as lack of comfort, energy efficiency and even the increase on the risk of accident [22]. The aforementioned behaviours are mainly influenced by the experience/inexperience of the drivers, as well as his/her age, gender and general health state [23]. Being these points considered, the assessment of DS appears as a powerful tool to identify the underlying causes of the aforementioned drawbacks.

Throughout bibliography, it can be seen that automated DS assessment has been performed to classify drivers among a variety of styles, cycles, and scenarios, regarding their behaviour and the road type [24–27]. This classification enables to consider the underlying causes of the deviations from the ideal driving behaviour and, consequently, to provide personalised recommendations. This advice might help drivers to mitigate the most undesirable characteristics of their handling, and consequently, achieve outstanding fuel consumption results [28], reduce their risk of accident or even increase the ride comfort [29]. It is worth to remark that, since these recommendations are intended to re-educate drivers if they follow incorrect driving patterns, they should not interfere with those that show correct ones.

To perform those personalised recommendations, machine learning techniques, such as fuzzy logic and artificial neural networks have been used to give coaching feedback to the driver about his/her performance regarding the DS, such as ref.[30], where DS-associated fuel consumption was assessed to provide reduction recommendations, or ref. [31], where critical manoeuvres are analysed to identify eco-driving-compromising points. In contrast, despite several works having been performed to estimate ride comfort [32] and passenger experience [33] by means of machine learning, there exists no proposals to provide DS recommendations to improve ride comfort.

In this work, we developed a method to improve ride comfort in an eco-driving framework. This approach, that could be used to develop advanced driving assistance system (ADAS), aims to reduce the DS-related discomfort in car occupants while taking into account eco-driving considerations. Thus, not only driving features that can trigger ride discomfort are identified, but also the eco-driving features are taken into account, providing personalised advice according to both fields. This corrects the eco-driving and wellness-compromising conditions. This ADAS was developed while using real-word data from an instrumented car, particularly the data stream from its CAN-bus and the inertial measurements unit (IMU) as well as simulated fuel consumption data.

We used self-organised maps (SOMs) [34, 35], a very popular unsupervised classifier in the fields of data mining and big data [36, 37], to characterise the DS. This intelligent, unsupervised machine learning algorithm is able to automatically group driving behaviours. It was chosen because it relies on a 2D representation of a high-dimensional complex system, known as a map, which is suitable for a qualitative evaluation of multiple driving behaviour features. Based on this characterisation, and taking into account its interpretation, some DS recommendations are set to be provided by the system. Moreover, this proposal has the main advantage of the aforementioned advice being provided by means of natural language, which makes it friendlier for the driver, always considering his/her individual characteristics. This fact is expected to encourage the engagement of the motorists with the system, who, otherwise, might not notice which points of their DS could cause a rise of the GHG emissions or the triggering of discomfort and motion sickness in their companions. Thus, in this work, we present the following contributions:

- A novel application of unsupervised neural networks to discover patterns that compromise eco-driving and ride comfort.
- Analysis an approach to the examination of the underlying causes of different types of non-optimal DSs for ride comfort from the eco-driving viewpoint.
- Personalisation of the provided advices when considering the aforementioned points. Those recommendations involve pedal, gear stick and steering wheel operation.
- Improvement in the ride comfort performance, with potential enhancements of up to the 57.7%.
- Improvement in the fuel-consumption ratios, with potential enhancements of up to the 47.1%.

These results stand out amongst those achieved by the traditional methods of gear recommendation [38] or eco-driving scoring [39] and have the additional advantage of facing the problematic putting the spotlight on the driver, instead of the vehicle. This approach, hence, treats both fuel consumption and ride comfort as a set of driver-dependent features, in contrast with the traditional solutions, which face the lack of comfort as a consequence of the fuel consumption–reduction techniques, such as in pulse-and-glide strategies, where the tradeoff between comfort and efficiency has to be carefully calibrated so as to not impairing the ride experience [40, 41]. Other works, such as [42] actually aboard the eco-driving from a behavioural viewpoint, but they achieve improvements of the 34.8%, which are clearly are superseded by ours.

The remainder of this paper is organised as follows. Section 2 introduces ride comfort and eco-driving concepts as well as the SOM modelling algorithm. Section 3 provides an overview of the proposed approach and describes the utilised instrumented car dataset. In addition, the most relevant features concerning eco-driving ride comfort are analysed and selected. In Section 4 the development of SOMs for joint ride comfort and eco-driving classification is provided. Different DS clusters are identified and natural language-based handling advice is developed. In Section 6 system validation and analysis is presented. Finally, Section 7 summarises the achieved improvements and exposes some final considerations.

## 2 | SOM-BASED RIDE COMFORT AND ECO-DRIVING CHARACTERISATION

Due to the main part of the proposed approach being based on the characterisation of DS taking into account both ride comfort and eco-driving through the use of unsupervised machine

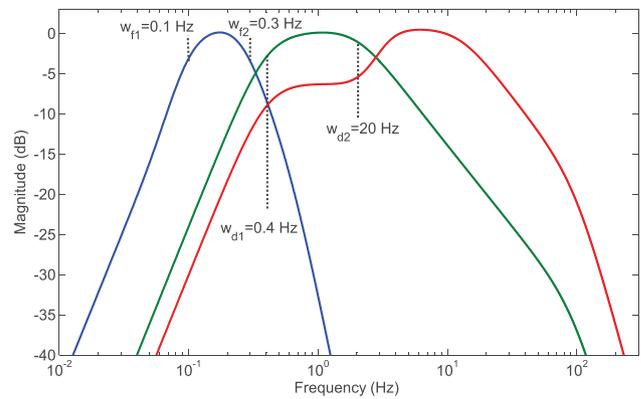

**FIGURE 1** Amplitude responses of different weighting filters in ISO 2631, $w_f$: motion sickness (blue), $w_d$: global comfort horizontal-component (green), and $w_k$: global comfort vertical-component (red)

learning algorithms, some basic theory on these concepts is introduced in this section.

### 2.1 | Ride comfort parameters

Two types of discomfort can be distinguished when we analyse the ride comfort during a given trip. In ref. [43] the general feeling of malaise is called average discomfort, while motion sickness is associated with dizziness, fatigue and nausea. The synergy between these two sensations causes the feeling of discomfort.

Two complementary types of approaches can be followed to assess the ride-quality experienced by passengers: qualitative and quantitative. Regarding qualitative methods, subjective tests [44] can be used to rate a variety of parameters from the viewpoint of the individual experience of the passenger. Conversely, several methods can be used to quantify the ride-quality during a given trip. In this line, the sensations caused by vibrations on the human body strongly depend on the signal direction and its spectral content. Hence, the International Standardisation Organisation (ISO) elaborated one of the mainly used standards: International Standard 2631 (ISO-2631-1) [45]. This standard describes ways to evaluate vibration exposure to the human body, defining methods to measure vibrations as well as how to process measurement data to standardised quantified performance measures concerning health, perception, comfort, and motion sickness.

In this standard, measurements are based on the frequency weighted root mean square (RMS) computations of acceleration data for each axis. This norm defines several filter shapes that delimit the frequency bands where different components of discomfort are present: filters $w_f$, $w_d$, and $w_k$, where filter $w_f$ is representative of motion sickness discomfort, while the filters $w_d$ and $w_k$ model the horizontal and vertical components of global discomfort, respectively.

As shown in Figure 1, the frequencies that mainly cause motion sickness are those between $w_{f_1} = 0.1$ Hz and $w_{f_2} = 0.3$ Hz (blue curve), so, the motion-sickness-associated measures have to be carried out for the input data filtered by the blue curve $w_f$. On the other hand, the green filter $w_d$ evaluates the sensation of general discomfort for a seated passenger when accelerations lie in longitudinal or lateral directions. Finally, the $w_k$ red filter is related with vertical accelerations.

Regarding the time persistent discomfort, the measured accelerations can be weighted and filtered in the way that ISO 2631-1 determines, where $a_{wxd}$, $a_{wyd}$, and $a_{wzk}$ are the results of being filtered by $w_f$, $w_d$, and $w_k$, respectively (see Figure 1). The weighted RMS acceleration for each axis is expressed as,

$$a_{wij} = \sqrt{\frac{1}{N}\sum_{k=1}^{N} a_i^2, w_j(k)}, \qquad (1)$$

where $i$ determines the direction, $w_j$ is the corresponding filter, and $N$ is the number of samples of the acceleration data.

Other significant parameter to be assessed is the likeliness of nausea by means of the computation of the motion sickness dose value (MSDV$_i$) for each axis by particularising the Equation (1) for $w_j = w_f$ (with $w_f$ being the motion sickness filter represented by the blue curve in Figure 1), such that,

$$\text{MSDV}_i = \sum_{k=1}^{N} a_i^2, w_f(k), \qquad (2)$$

which is a standardised measure of motion sickness.

There are several works related to the analysis of those parameters suggesting some variants. Concerning motion sickness, although International Standard pays attention to vertical accelerations, there are later works that prove the influence of lateral accelerations in motion sickness. Thus, in ref. [46], the likeliness of nausea is represented by the Equation (2) applied to lateral accelerations. Additionally, in ref. [47], the ISO weighting filters shown in Figure 1 are slightly modified to better match with the real sensations caused by the transverse forces, being $w_{f1} = 0.02$ Hz and $w_{f2} = 0.3$ Hz. On the other hand, in ref. [48], the probability of a car occupant to get motion sick enough to vomit is suggested, as well as several weighting parameters, both for vertical and horizontal accelerations.

In this work we chose the filter proposed in ref. [47], and we analysed motion sickness parameter using lateral accelerations, since they depend more on driving than the vertical ones. In addition, based on ref. [48], we combined the contribution of both vertical and horizontal accelerations by means of the new vomit rate parameter presented in Equation (3). Moreover, despite those parameters being accumulative (depends on travel length), in this work a window-based averaging was used.

$$\text{VR} = \sqrt{\left(\frac{1}{3}\right)^2 \text{MSDV}_z^2 + \left(\frac{\sqrt{2}}{3}\right)^2 \text{MSDV}_y^2}. \qquad (3)$$

Finally, acceleration and jerk peaks are evaluated using a methodology based on acceleration thresholds [49]. A high value of acceleration or jerk can cause discomfort even during shorter periods of time. When the levels get too high the passenger will find it difficult to maintain posture. Limit values vary between the studies. In ref. [50] a maximum acceleration value of 1.47 m s$^{-2}$ is determined whereas in ref. [43] it is argued that since an automobile only carries seated passengers it is expected that the thresholds should be set on the higher side than in a train or on a bus and the limit is set closer to 2 m s$^{-2}$.

Thus, we can assess the transient discomfort, by counting the acceleration peaks with values above a certain threshold, such that:

$$n_i = n_i + 1 \text{ when } a_i > \text{threshold} \qquad (4)$$

where $i$ determines the direction of the acceleration in each of the XYZ axes and the threshold was fixed at 1.75 m s$^{-2}$.

## 2.2 | Eco-driving and ride comfort characterisation

Several approaches can be followed to alleviate and improve the scenario of high emission levels, with the undersizing of the engines to reduce both fuel consumption and, consequently, GHG, being the most popular measure [51]. However, while this technique has been applied for more than 20 years by increasing the influence of electronics on the control of the engine, it should be remarked that this technique is less effective for compression ignition engines, being a topic of major concern particularly in Europe because it holds the largest ratio of diesel vehicles [52]. Additionally, regulations force manufacturers to deploy automated fuel economy-intended systems in production automobiles. These systems, such as gear recommendation [38] or eco-driving scoring [39], despite achieving the objective of reducing the emissions of pollutants [38], could be further enhanced by means of a holistic solution which combined the behaviour of the drivers, a previous training on eco-driving techniques and the efforts of car manufacturers to effectively achieve a significant level of improvement.

Several authors remark that eco-driving could effectively contribute to reducing overall fuel consumption and GHG emissions if adequate education about strategic, tactical, and operational decisions were provided to drivers [53–55], so, the use of several techniques, can help drivers to maximise fuel efficiency and to reduce pollutants' emissions. In a broader sense, eco-driving factors could be categorised into three main groups: strategic decisions (vehicle selection and maintenance), tactical decisions (route planning) [41], and operational decisions (driving style) [56]. In this study, eco-driving focuses on the most useful eco-driving skills that every driver can put into practice, independently of strategic and tactical decisions [57]. The driving style involves driving speed, acceleration, deceleration, gear changing, idling, and accessories (i.e. air conditioning). In ref. [56], the authors compare the ranges of percentages of fuel savings or $CO_2$ reduction contributed by each of the above eco-driving factor. They concluded that the primary eco-driving factor is acceleration/deceleration, contributing to 3.5–40% fuel savings. Driving speed could contribute to 2–29%, while idling reduction accounts for 6–20%. Other factors, such as air conditioning, despite having an effect on fuel consumption are determinant in passengers' comfort. Moreover, several safe driving techniques or "golden rules of eco-driving" have been developed [58]:

- Avoid unnecessary acceleration and braking and make maximum use of the vehicle's momentum.
- Maintain a steady speed at low RPM: drive smoothly, using the highest possible gear at low RPM.
- Shift up early: shift to higher gear by approximately 2,000 RPM.

Other useful driving advice which can lead to fuel savings are: check tire pressures frequently, remember all accessory loads that add to fuel consumption, use electrical equipment sparingly, and avoid carrying dead weight and adding unnecessarily to aerodynamic drag.

However, in some scenarios, the above eco-driving rules do not consider the ride comfort viewpoint. As an example, in ref. [40] the tradeoff between ride comfort and fuel efficiency are studied for the pulse and glide strategies. These strategies require the clutch to be disengaged between engine pulses to ride the car by inertia while the traction chain is not linked to the wheels. However, this may severely condition the passengers' comfort, since high values of vibration and jerk can happen, and, consequently, several levels of calibration have to be applied to not disturbing the occupants [40]. In the same fashion, some sources propose minimising the required time to achieve the speed setpoint so as to minimise the fuel consumption [59, 60], which returns outstanding economy rates by the cost of conditioning the passenger comfort in both the danger sense and sensation terms (i.e. jerkiness, noise, accelerations etc.). For those reasons, a tradeoff solution has to be found to guarantee both the fuel economy and the ride comfort for the vehicle occupants.

To better analyse the aforementioned tradeoff, among many of the previously mentioned methods, SOMs have been successfully used to online cluster DS regarding meaningful features and to provide advice to modify the undesirable aspects of car handling [30].

## 2.3 | Self-organising map clustering algorithm

The SOM is an unsupervised artificial neural network suitable for clustering and visualisation of complex multi-dimensional data [34, 35, 61]. It defines a mapping from a set of high-dimensional input data onto a regular low-dimensional discrete grid, known as feature map. The SOM algorithm, presented in ref. [35], is based on a competitive learning algorithm, the winner-take-all. In the winner-take-all algorithm an input vector is represented by the closest neuron prototype vector, which is assigned during training to a data cluster centre. These

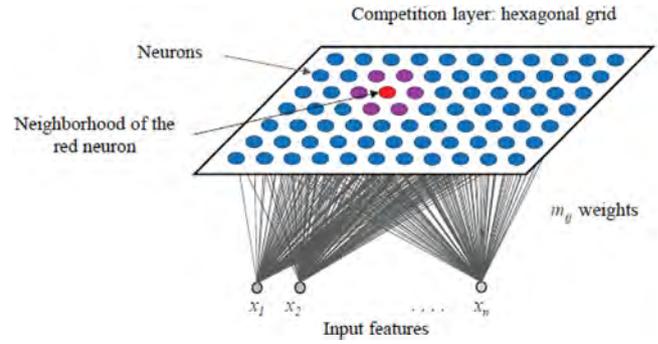

**FIGURE 2** Typical SOM topology: structure of an $N$-input SOM, $\mathbf{x} = (x_1, x_2, \ldots, x_N)$, and $M=77$ output neurons distributed into a 7×11 2D hexagonal grid

prototype vectors are stored in the weights of the neural network.

### 2.3.1 | Topology of self-organising maps

The architecture of a SOM consists of a topologically organised array of neurons set along a regular grid: the competition layer. Each input to the SOM is fully connected to every neuron in the competition layer. Figure 2 depicts a 2D output layers with $M = 77$ neurons set along an hexagonal grid.

Each neuron in the competition layer has a double representation: an $N$-dimensional vector $\mathbf{m}_i = (m_{i1}, m_{i2}, \ldots, m_{iN})$, $1 \leq i \leq M$, known as weight vector, and its position in the grid. The number of components of the vector is equal to the number of input features $N$.

Clustering a dataset by means of SOM paradigm is carried out using a two-level approach: first the SOM is trained, and then, the prototype vectors of the SOM are clustered [62].

### 2.3.2 | Training and clustering of self-organising maps

First, an initial weight is randomly assigned to each neuron connection [63]. Then, in each training step, one input sample $\mathbf{x}^k = (x_1^k, x_2^k, \ldots, x_N^k)$, $1 \leq k \leq K$, from the training set is chosen randomly and the distances between this sample and all the neuron weights of the SOM are computed. The Euclidean distance is most frequently used as the distance metric $\| \cdot \|$.

$$\| \mathbf{x}^k - \mathbf{m}_i \|^2 = \sum_{j=1}^{N} (x_j^k - m_{ij})^2. \qquad (5)$$

The output neuron whose weight vector is nearest to the $k$th input sample, according to Equation 5, is the best matching unit or the winner neuron, usually denoted by $c$. The best matching unit is used to update the weight vectors of the SOM. In this process, the best matching unit and its neighbours are moved towards the $k$th input sample, bringing them closer. For each neuron of the SOM, the weight vector is updated as follows,

$$\mathbf{m}_i(n+1) = \mathbf{m}_i(n) + \alpha(n) h_{ci}(n) \|\mathbf{x}^k(n) - \mathbf{m}_i(n)\|, \quad (6)$$

where $n$ denotes the iteration step, $\mathbf{x}^k(n)$ is an input sample randomly selected from the training set at iteration $n$, $h_{ci}(n)$ is a neighbourhood function or kernel around the best matching unit and $\alpha(n)$ is the learning rate. Both $\alpha(n)$ and $h_{ci}(n)$ are decreasing functions approaching zero with each iteration. The neighbourhood function specifies how much the $i$th neuron has to move toward the input sample at iteration step $n$. It is a radial basis function, usually a Gaussian function centred at the best matching unit.

The unified distance matrix (i.e. U-matrix), composed of the distances of the weight vectors to each of their neighbours in the grid, provides an insight into the SOM response. It is not only a powerful analysis tool, suitable for identifying clusters mathematically, but also a useful visualisation tool.

In sum, after initialisation, the following training steps are repeated until a stop criterion is reached:

1. Competition. Given a randomly selected input sample, all the neurons in the competition layer compete with each other to be the best matching unit. The neuron that is closer to the input sample is the winner (i.e. the winner-take-all).
2. Cooperation. The best matching unit also excites the neurons in its neighbourhood. This cooperative process decays as neurons are further away from the winning neuron.
3. Adaptation. The best matching unit and its neighbouring neurons are pulled closer to the input sample. For each neuron in the SOM, the weight vector is updated accordingly to Equation 6.

The trained SOM provides a nonlinear mapping of the multidimensional dataset onto a 2D grid that allows identifying groups of samples with similar characteristics (i.e. clusters) by taking into account all features simultaneously. A clustering algorithm based on the U-matrix and the k-nearest neighbours algorithm (k-NN) was used [64] with the aim of identifying and assessing meaningful clusters. In this work, the training procedure will be carried out by using the Matlab Neural Network Clustering App [65] while the clustering development will be performed by means of the CIS SOM Toolbox for Matlab [64].

## 3 | SYSTEM OVERVIEW AND DRIVING STYLE FEATURES

To achieve our objective of developing an ADAS to improve ride comfort while paying attention to eco-driving, we performed an in-depth analysis of the DS of 20 drivers of different age-groups and driving experience levels. We used data collected from the real-world using an instrumented car. This information was used to discover underlying DS characteristics, which are to be decoded from the raw data by means of the unsupervised SOM clustering algorithm.

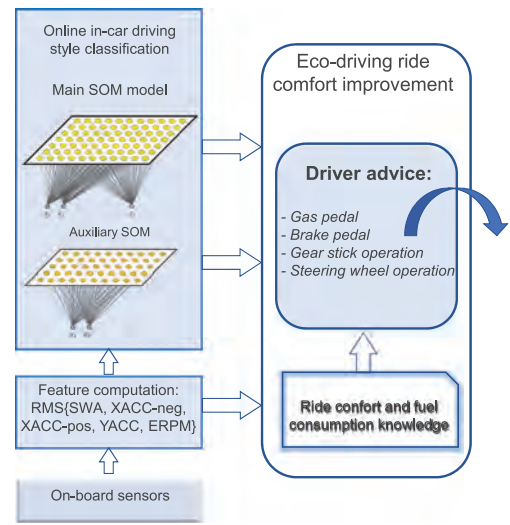

**FIGURE 3** Block diagram of the proposed ride comfort and eco-driving assessment system

Given a set of driving signal measurement, the devised model is able to perform online classification and to provide the driver with personalised driving advice in real time, with the aim of improving his/her global ride comfort and eco-friendliness. Figure 3 shows a block diagram of the eco-driving ride comfort ADAS. The system is composed of a feature computation block, a SOM-based driving style classifier, and the driver advice module.

The development of the proposed ADAS for ride comfort improvement was performed in three stages. First, the selection of a set of meaningful features, able to account for both ride comfort and eco-driving viewpoints, was performed by means of analysing the magnitude of their correlation coefficients. Next, the SOM unsupervised clustering technique was applied to discover different driving styles. Two SOMs were trained, the first one (i.e. main SOM) is able to classify the driving data into a number of classes that account fundamentally for ride comfort, while the second SOM (i.e. auxiliary SOM) deals with fuel consumption features. After a quantitative and qualitative evaluation of each SOM, well differentiated DS clusters were identified and labelled according to their ride comfort characteristics and eco-driving measures. After that, with these pieces of information, particular actions on the car controls (i.e. gas pedal, brake pedal, steering wheel, and gear stick) were developed in order to provide advice to the drivers.

### 3.1 | Dataset

A variety of studies on intelligent vehicles have been performed with the aim of discovering the way how vehicles, traffic, motorists, and their surrounding environment relate and behave, as well as to select the best sensors to register each situation correctly. Most driving studies rely on dedicated instrumented cars, fitted with several sensors jointly used with loggers of the field buses of the vehicle, instead of participants riding their own vehicles, to simplify data collection and delivery flow. This kind of studies (e.g. NU-Drive [66], UTDrive [67], or Uyanik [68]) provide more consistent and reproducible data records, produced under a more strict experimental control, allowing researchers to establish comparisons between different drivers more easily than if they had driven their own cars.

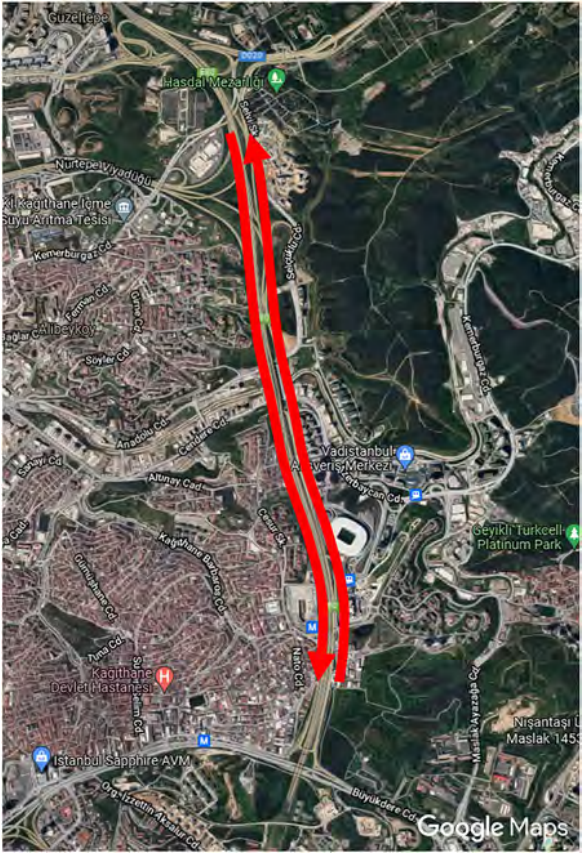

**FIGURE 4** Stretches of the Uyanik route used in this work. An average of 9:27 min of driving is available for each driver

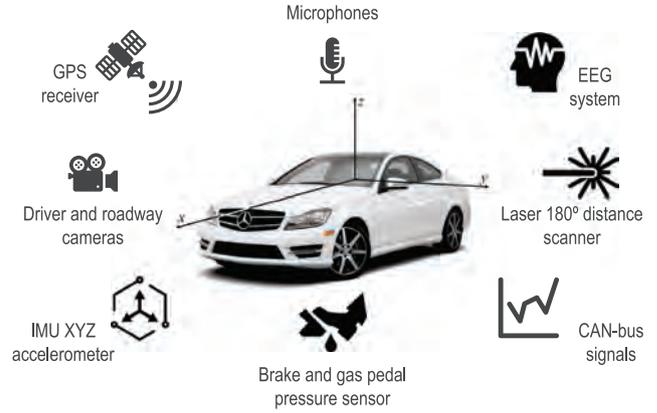

**FIGURE 5** Data-acquisition systems and sensors installed in the instrumented car

In this work, we used the Uyanik dataset [68], from the University of Sabançi at Istanbul. This study's car collects three channels of uncompressed video from the left and right sides of the driver and the road ahead (see Figure 5). It also includes three audio recordings, GPS, and CAN-bus readings, including vehicle speed (VS), engine RPM (ERPM), steering wheel angle (SWA), and brake pedal status (pressed or idle). Gas pedal engagement percent (PGP) is sampled at either 10 or 32 Hz, whereas brake pedal and gas pedal pressure sensor readings (BP and GP, respectively) are sampled at the same CAN-bus rate. Finally, a laser distance measuring device was fitted in the front bumper jointly with a three-axis XYZ inertial IMU set-up. In this study, all of the signals were handled jointly, which requires a re-sampling of the data streams to the highest frequency of 32 Hz and their proper synchronisation. This re-synchronisation was carried out by displaying the video feeds jointly with the plain data of Uyanik, resulting in a set of spreadsheet-like data chunks that can be easily processed automatically.

The driving behaviour data collection was performed in Istanbul. The car route is around 25 km (about 40 min), and includes different kinds of sections: city, highway, secondary roads, and a university campus. The age range for female drivers was 21–48, and the corresponding male range was 22–61. The route was the same for all 20 drivers, however, the road conditions differ depending on traffic and weather. In this work, we selected the stretches of route remarked in Figure 4, which exclusively comprehend highway-type roads. While riding along these roads, which show fluent traffic, vehicles acquire high mean speeds with low deviations. Additionally, these highway stretches were selected so that their mean slopes were lower than the 2%, so drivers had to uninterruptedly operate the gas pedal to adjust their vehicles' speeds to the traffic flow.

To complete the dataset so as to be able to assess the fuel consumption and, consequently, the eco-driving characteristics, we added the fuel consumption data corresponding with each sample. This data, which the original dataset lacked, was obtained by introducing the real-world driving data into a realistic model of the instrumented car [30], developed with the GT-Suite simulator [69]. This model, as detailed in ref. [30], takes into account parameters of the car (e.g. car wheelbase, wheel radius, friction coefficients, aerodynamics, weight, inertia, and final transmission ratio), the individual ratios for each of the user-selectable gears, the engine parameters and, most importantly, the telemetry data related to the handling of the car, such as selected gear, accelerator pedal state, brake pedal state, clutch, and desired speed.

## 3.2 | Statistical analysis of driving behaviour

With the aim of comparing the driving style of individual drivers concerning ride comfort and eco-driving, a statistical analysis of comfort parameters and fuel consumption was performed. Figure 6 depicts mean parameter values corresponding to each driver while completing the same route. As can be seen, mean fuel consumption differs among drivers, ranging from 2.09 $l/100$ km (driver 6) to 4.17 $l/100$ km (driver 18). Moreover, it can be highlighted that drivers 6, 11, 12, and 20 consume less than 3 $l/100$ km, while drivers 14 and 18 are slightly above 4 $l/100$ km, being the remainder drivers between those values. The first group of drivers, namely eco-friendly drivers, exhibit low positive and negative acceleration peaks in $x$ axis (i.e. $n_x$ pos and $n_x$ neg), as well as relatively low $y$ axis acceleration peaks (i.e. $n_y$). These drivers show also low motion sickness dose values and vomit rates. For example, average vomit rates are: 0.79 m s$^{-2}$ (Driver 6), 0.76 m s$^{-2}$ (Driver 11), and 0.82 m s$^{-2}$ (drivers 12 and 20). In consequence, it can be concluded that the most ecological drivers from fuel consumption viewpoint present also a comfortable driving style. On the contrary, the drivers that consume the most fuel are not necessarily the most uncomfortable ones (e.g. driver 18 shows high fuel consumption: 4.15 $l/100$ km and low vomit rate: 0.86 m s$^{-2}$). Thus, it can be seen that there is a certain relationship between driving comfort and fuel consumption, but this synergy must be analysed in depth.

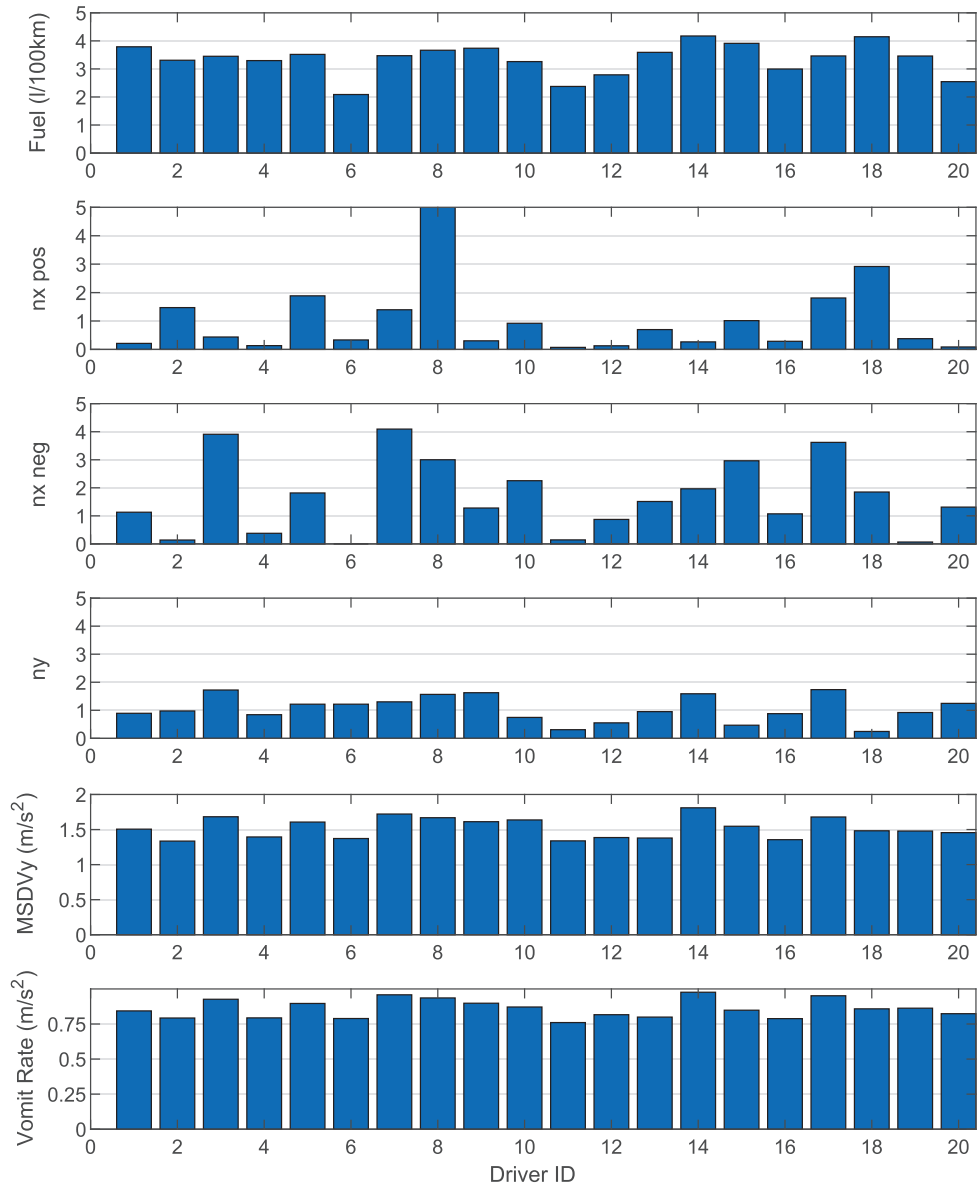

**FIGURE 6** Average ride comfort parameters and fuel consumption for each driver. These parameters are computed on a sample of 2290 driving windows

Next, a different statistical approach to fuel consumption and comfort parameters is provided. The kernel density estimation is used to shed light on the nonlinear relationship between the driving behaviour of selected drivers and these parameters. The kernel density estimation technique, unlike histogram, produces smooth estimate of the probability density function and is able to suggest multimodality [70]. It is useful to estimate the probability density function of datasets difficult to be modelled by parametric density functions.

Figure 7 depicts the bivariate kernel density function of fuel consumption and vomit rate corresponding to six representative drivers that exhibit different statistical driving behaviour. The probability density function shows different shapes by each one of the selected drivers. The positions of the surface peaks indicate maximum likelihood of fuel consumption and vomit rate. Furthermore, the sharper the surface, the more regular the driving style. To carry out this selection, the kernel density function for each driver was elaborated, and consequently, two main trends were identified between all the drivers. As can be seen, both drivers 6 and 20 present a rather regular and comfortable driving style, while driver 6 peaks at the lowest values both in fuel and vomit rate. In contrast the peak values corresponding to driver 20 are slightly greater than the former ones. However, concerning fuel consumption, driver 6 shows a more disperse trend than driver 20. Opposite, the coordinates of the peak for drivers 8, 14, and 18 show high average fuel consumption and irregular driving patterns (i.e. flattened surfaces). Moreover, driver 18 exhibits a clear bimodal driving style. It is worth noting that although the secondary peak is unlikely, it increases the average fuel consumption. These results are consistent with those shown in Figure 6. In sum, the driving data construct different kernel density estimation surfaces according to the drivers' preferred driving style. Some drivers are more regular than others, although external factors such as weather or road conditions, could also be in part responsible of these differences.

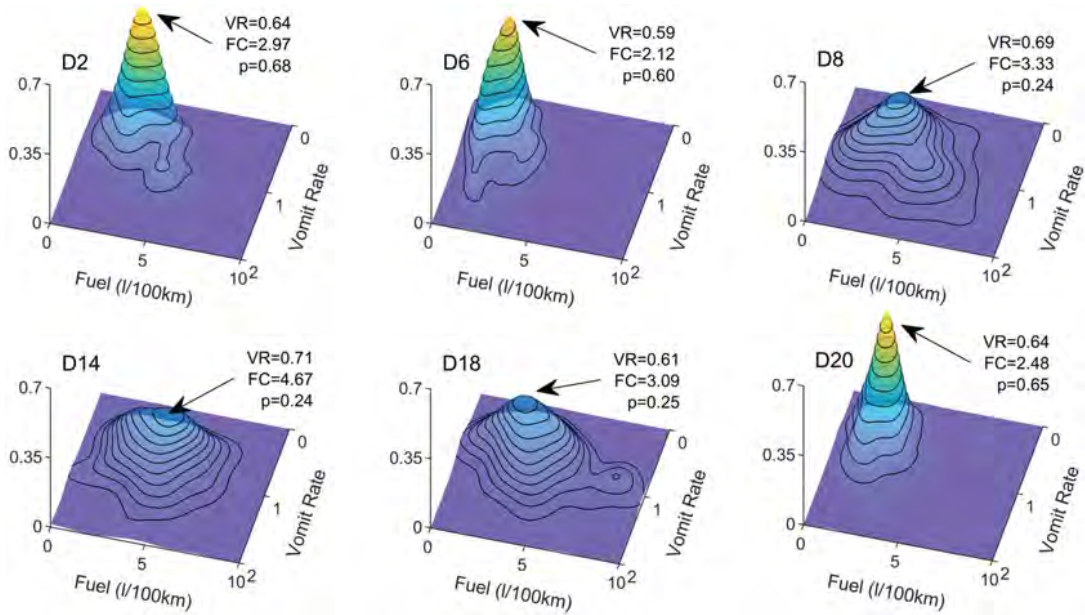

**FIGURE 7** 3D kernel density estimation, computed on a total of 2290 driving windows, links vomit rate and fuel consumption for some of the most representative drivers of the sample population. Wider projections onto the horizontal plane mean higher variances of the vomit rate and the fuel consumption measured for the driver

## 3.3 | Feature selection

With the aim of choosing DS variables with a significant level of relationship with both the comfort parameters and fuel consumption introduced in Section 2, we performed a correlation analysis with the real-world driving signals. The features were computed over 256-sample windows (i.e. 8 s at a 32 Hz sample rate) with 50% overlapping between consecutive windows (i.e. 128 samples, or 4 s). It is worth noting that the most fuel-demanding sections of the instrumented car route were selected, that is to say, those that ran through highway and motorway. Moreover, sections with traffic jams and slow traffic (i.e. mean speed below 60 km h$^{-1}$) were discarded with the aim of avoiding outliers.

Table 1 summarises the Pearson correlation coefficients (PCC) of the set of selected driving characteristics with different discomfort parameters and fuel consumption. The first column of each variable represents the root mean square (RMS) of the signal whereas the second represents the variance. As can be seen, the features with the strongest correlations with fuel consumption and the selected discomfort measures are highlighted. For those features concerning the X-axis of the car (longitudinal direction, see Figure 5), we can see that both the RMS and variance of XACC pos and XACC neg signals present strong correlation with nx pos and nx neg, respectively. Moreover, the variance of VS and both RMS and variance of XACC correlate with nx neg positively. In the same way, for those features concerning the Y-axis of the car (transverse direction, see Figure 5), both RMS and variance of YACC and SWA are strongly correlated with MSDV$_y$ and VR. Finally, regarding fuel consumption, mean and variance of XACC pos and RMS of ERPM are the most informative features. It is worth noting that the Z-axis features, despite having a noticeable contribution on worsening the motion sickness felt by the passengers, were discarded since they do not directly depend on the DS, but on the road characteristics.

In sum, the results displayed in Table 1 show that the features with the strongest correlation with ride comfort are: RMS{SWA, XACC, XACC neg, XACC pos, YACC} and

**TABLE 1** Pearson correlation coefficients (PCCs) between eco-driving ride-comfort variables and real-world data from the instrumented car. RMS and variances are computed using 2290, 8 s analysis windows

| | SWA RMS | SWA Var | VS RMS | VS Var | XACC RMS | XACC Var | XACC neg RMS | XACC neg Var | XACC pos RMS | XACC pos Var | YACC RMS | YACC Var | ERPM RMS | ERPM Var |
|---|---|---|---|---|---|---|---|---|---|---|---|---|---|---|
| Fuel[*l*/100 km] | 0.07 | 0.05 | 0.26 | 0.19 | 0.16 | 0.25 | −0.35 | −0.14 | **0.60** | **0.50** | 0.05 | 0.08 | **0.45** | 0.15 |
| nx pos | 0.12 | 0.15 | −0.07 | 0.18 | 0.33 | 0.37 | −0.08 | −0.01 | **0.45** | **0.61** | 0.16 | 0.20 | 0.07 | 0.16 |
| nx neg | 0.09 | 0.10 | −0.01 | **0.49** | **0.61** | **0.53** | **0.65** | **0.79** | −0.10 | −0.04 | 0.09 | 0.10 | −0.08 | 0.20 |
| ny | 0.28 | 0.28 | 0.12 | 0.03 | 0.11 | 0.18 | 0.06 | 0.13 | 0.09 | 0.11 | **0.65** | **0.70** | 0.02 | −0.02 |
| $MSDV_y$[m s$^{-2}$] | **0.68** | **0.60** | 0.05 | 0.21 | 0.19 | 0.21 | 0.05 | 0.14 | 0.16 | 0.16 | **0.73** | **0.60** | 0.02 | 0.07 |
| VR[m s$^{-2}$] | **0.63** | **0.56** | 0.09 | 0.20 | 0.20 | 0.23 | 0.04 | 0.14 | 0.19 | 0.19 | **0.76** | **0.65** | 0.02 | 0.07 |

variance{SWA, VS, XACC, XACC neg, XACC pos, YACC}. On the other hand, fuel consumption is positively correlated with RMS{XACC pos, ERPM} and variance{XACC pos}.

## 4 | DEVELOPMENT OF THE SOM-BASED CLASSIFIER

According to the scheme depicted in Figure 3, an unsupervised SOM clustering algorithm was used to classify drivers regarding their ride comfort and eco-driving ratings. With this purpose, we selected the entire set of samples which comprises driving data of 20 drivers. As an average of 115 windows is available per individual, the whole set of driving samples consists of 2290 windows (i.e. more than 2.5 driving hours). The 75% of the 5D samples will be used to train a SOM, which is to say $K = 1717$, keeping the remaining quarter for testing purposes. The data were normalised before training in order to avoid distortion in the results due to the use of Euclidean distances (Equation (5)).

A comprehensive series of training experiments revealed that a reduced subset of only five independent features is able to model the jointly relationship ride-comfort/fuel-consumption with driving signals in a very satisfactory way. These features are: RMS{SWA, XACC neg, XACC pos, YACC, ERPM}. In contrast, despite a number of variances show high correlation coefficients, they were found less suitable for modelling driving behaviour than the corresponding RMS values. The above selected features will be used to develop a SOM-based driving style model and provide drivers with specific recommendations.

The number of output neurons of the SOM was initially selected with the Vesanto's rule [64], which defines the optimal number of neurons as $M = 5\sqrt{K}$. Thus, a 14 × 14 SOM topology (i.e. $M = 196$) was defined and recursively trained. However, we carried out experiments ranging from 13 × 13 to 15 × 15 maps, achieving the best results for the latter, so, a 15 × 15 map was built and trained to improve the feature extraction capabilities. In Figure 8 the SOM results are analysed by means of displaying the input weight planes. These planes depict the weights associated with each input for each neuron, and reflect the input magnitudes that are expected to cause a hit for each neuron (i.e. the neuron is the best matching unit for that particular input). Thus, considering that lighter colours represent the higher values, several assertions can be made by visually analysing the weight values. As can be seen, all the five planes are clearly different, which indicates that the inputs show no correlation between them. This fact proves that the provided information has no redundancies, improving the separation capabilities of the SOM.

It can be seen that drivers with high values of SWA tend to show also high levels of YACC, fact that conditions the ride comfort of the car occupants (see top left corner). On the other hand, several neurons of the bottom left part of both ERPM and XACC pos show high values as well, which indicates that not only are those samples related to a high fuel consumption, but also might show a relationship with discomfort. Regarding XACC neg, high values for this variable are displayed at the top-right part of its corresponding map, which could show a certain level of relationship with negative acceleration peaks, derived of spurious braking. Finally, it is worth noting that the weight map for ERPM is particularly uniform, which might indicate that the extraction of fuel consumption characteristics for this specific SOM could be less conclusive than for the other features.

To extract more solid conclusions that those provided by the visual inspection of the input planes, several partitions can be made on the map in order to exploit the granularity of the algorithm and to group similar neurons into defined clusters. It should be remarked that, according to the k-NN algorithm referred to in Section 2.3, generally, the finer the partition, the greater the number of clusters found in the SOM. In this case, a three-cluster partition was used as an starting point to assess the relationship with both the ride comfort and the fuel consumption of drivers. This partition is shown in Figure 9.

When comparing Figure 9 with Figure 8, we can see several similarities, particularly with the input maps of SWA and YACC. Thus, if we compare again the top left section of the input map of SWA, we can see that the light neurons surrounded by the dark ones of the upper left corner match with the red neurons of the upper left corner of Figure 9. The same happens with the bottom left corner of YACC, for which, when comparing with the bottom left corner of Figure 9, it can be seen that the lighter neurons correspond with the elements of the red cluster, while the darkest ones correspond with the blue coloured elements.

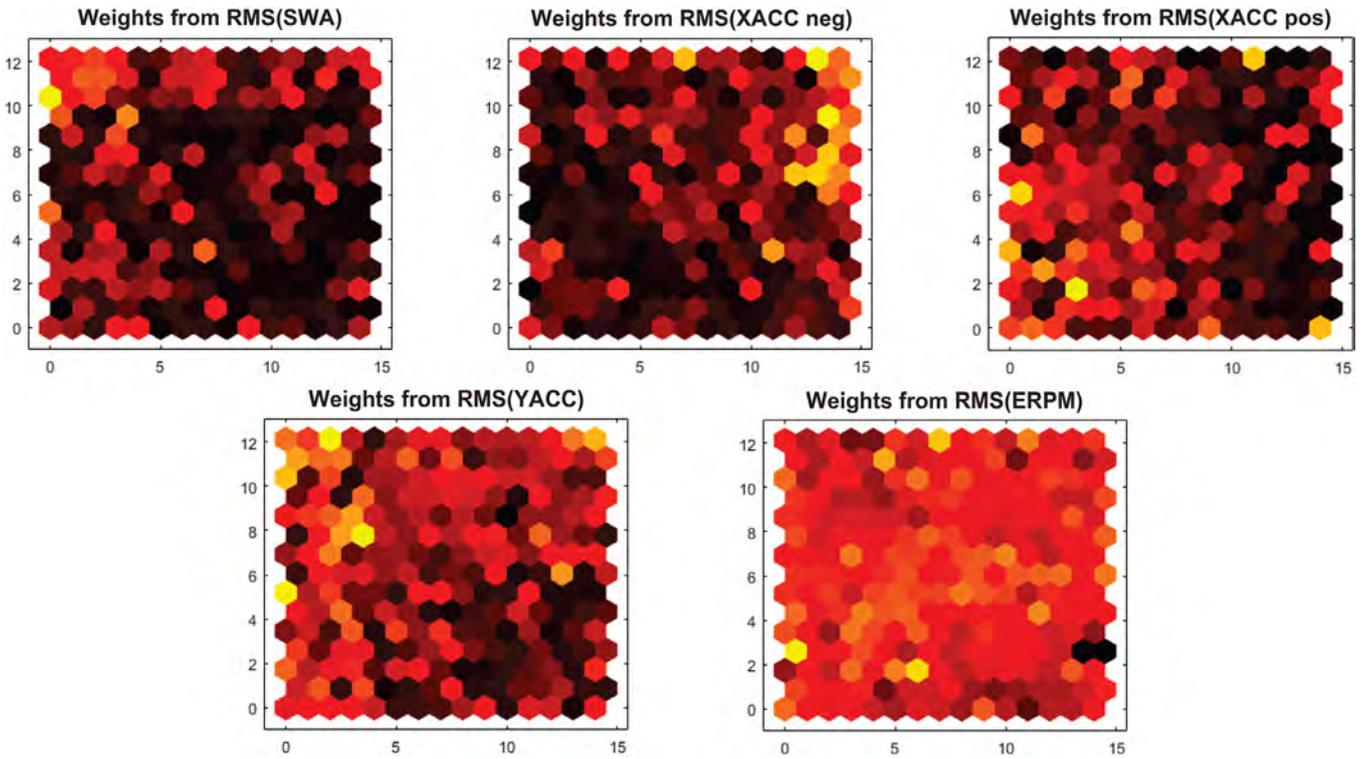

**FIGURE 8** Input weight maps for the five selected input features, RMS{SWA,XACC neg,XACC pos} and RMS{YACC,ERPM}, from left to right and top to bottom. A total of 1717 data windows were analysed. Lighter colours represent higher values of the corresponding features for each of the 15 × 15 neurons of the SOM. These values are used to later partition the DS clusters because they depict the weights associated with each input for each neuron, and reflect the input magnitudes that are expected to cause a hit for each neuron

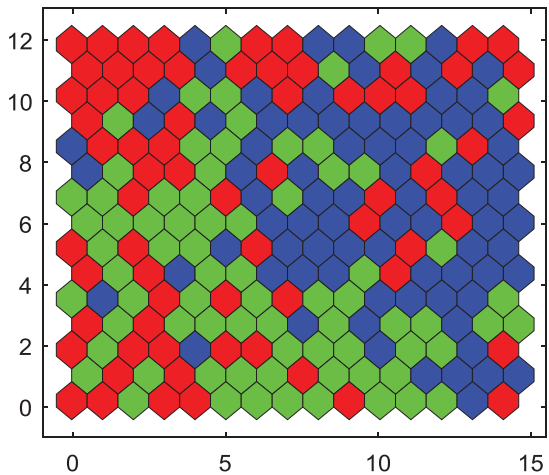

**FIGURE 9** SOM partitioned to separate neurons between three clusters of ride comfort and fuel consumption. This partition is based on the weights obtained during the training of the SOM and produced by carefully selecting the $k$ of the k-NN partitioning algorithm

**TABLE 2** Average values and variances for discomfort and fuel consumption for the three-cluster classification. This classification is performed on the total of 2290 windows

| Cluster/variable | | Blue | Green | Red |
| --- | --- | --- | --- | --- |
| $MSDV_y$ | Avg. | 1.27 | 1.36 | 3.00 |
| [m s$^{-2}$] | Var. | 0.31 | 0.33 | 1.24 |
| Vomit rate | Avg. | 0.73 | 0.81 | 1.50 |
| [m s$^{-2}$] | Var. | 0.07 | 0.07 | 0.27 |
| nx pos | Avg. | 0.05 | 1.40 | 2.73 |
| | Var. | 0.19 | 36.8 | 94.2 |
| nx neg | Avg. | 2.30 | 0.42 | 3.69 |
| | Var. | 76.7 | 11.3 | 175.7 |
| ny | Avg. | 0.42 | 0.68 | 4.78 |
| | Var. | 1.95 | 3.87 | 48.0 |
| Fuel cons. | Avg. | 2.92 | 3.64 | 3.68 |
| [$l$/100 km] | Var. | 1.98 | 2.71 | 4.95 |

Once the three different clusters are displayed and analysed, a quantitative evaluation of the ride comfort and fuel consumption parameters is performed. For that purpose, the mean values and variances of the selected variables are computed and displayed in Table 2. As can be seen, the red cluster, which according to the description of the input planes has high values of SWA and YACC, depicts average values of the continuous discomfort variables that double those for the green and blue clusters, which compile much lower SWA and YACC values. In the same fashion, Table 2 shows that nx pos and ny, which model transient discomfort peaks, are noticeably higher for the red cluster than for the blue one. This trend, however, is not kept for nx neg, which has a minimum for the green cluster, increasing for the blue one. Regarding variances, a general increasing trend of variance jointly with the average values can be seen. This trend is related to the kernel density estimation graphs on Figure 7, where low consumption, low motion sickness drivers show more pointy surfaces with their probabilities lying in a much smaller area, reflected in the low variance of the low motion sickness drivers. Thus, in the case of the motion sickness variables, their variances are relatively low, which means that the generated clusters are fairly compact and well separated for the assessment of this feature. In contrast, the transient discomfort peaks' variance values are high due to some drivers being more likely to present spurious acceleration peaks above a certain threshold, even for the same class, as displayed in Figure 6.

**TABLE 3** Average values and variances for discomfort and fuel consumption for the five-cluster classification. This classification is performed on the total of 2290 windows

| Cluster/variable | | Blue | Green | Yellow | Magen. | Red |
|---|---|---|---|---|---|---|
| $MSDV_y$ | Avg. | 1.09 | 1.19 | 1.65 | 1.85 | 3.25 |
| [m s$^{-2}$] | Var. | 0.18 | 0.22 | 0.56 | 0.46 | 1.58 |
| Vomit rate | Avg. | 0.64 | 0.72 | 0.94 | 1.01 | 1.62 |
| [m s$^{-2}$] | Var. | 0.04 | 0.05 | 0.11 | 0.09 | 0.34 |
| nx pos | Avg. | 0.02 | 0.18 | 3.70 | 0.39 | 0.98 |
| | Var. | 0.05 | 0.78 | 116.3 | 1.84 | 17.3 |
| nx neg | Avg. | 1.20 | 0.21 | 0.37 | 5.27 | 2.21 |
| | Var. | 26.9 | 1.87 | 7.18 | 229.8 | 61.6 |
| ny | Avg. | 0.05 | 0.23 | 1.02 | 2.02 | 5.74 |
| | Var. | 0.16 | 0.89 | 5.43 | 10.5 | 69.2 |
| Fuel cons. | Avg. | 2.82 | 2.86 | 4.90 | 3.04 | 3.08 |
| [l/100 km] | Var. | 1.70 | 1.25 | 2.94 | 2.44 | 3.45 |

It is worth to remark that the average values of comfort variables are slightly lower for the blue cluster than for the green one, being the difference between the average values of fuel consumption the major determinant for their separation. Conversely, for fuel consumption, the identified cluster do not show such that level of compactness and separation, since the average values are similar for both red and green clusters, while the variances are too high, being, in the case of the red cluster, even higher than the mean value. For these reasons, this partition can be considered valid for ride comfort, but not conclusive for fuel consumption, so, the fineness of the clustering was further increased until a five-class grouping was achieved.

To assess how the finer clustering affects to the discomfort and fuel consumption ratings, the average and variance values were extracted and displayed in Table 3. In this case, the red cluster stands out among the remaining four clusters in terms of vomit rate and $MSDV_y$, which are, by far, the highest. With respect to the other clusters, it can be seen that a full spectrum of classes, ranging from intermediate to low vomit rate and $MSDV_y$, has been found. Regarding the discontinuous discomfort variables, the trend displayed for three-clusters is kept.

Thus, from the maximum displayed for the magenta class, nx neg uniformly decreases jointly with vomit rate and $MSDV_y$ until the blue cluster is reached, increasing. As for fuel consumption, it can be clearly seen that the samples with the high to intermediate continuous discomfort values do not match with the highest fuel consumption rates. On the other hand, the samples with the lowest vomit rate are those showing the lowest fuel consumption rates as well. Nevertheless, the yellow cluster stands out among the others, since, despite having an intermediate level of vomit rate, it shows a very high fuel consumption level, jointly with an increase on nx pos. This means that the continuous ride discomfort is not always related to the eco-unfriendliness of drivers, while the spurious might be, as stated in Section 3.2.

Nonetheless, when inspected 2-by-2, the addressed clusters, particularly the couple blue-green and the couple yellow-magenta are very similar in terms of ride comfort. This resembles a three-cluster classification, so, instead of using the latter five-class clustering, the former three-class one seems simpler and useful to extract conclusions about ride comfort.

## 4.1 | Auxiliary eco-driving SOM

Taking the results displayed in the previous section, an auxiliary, fuel consumption-dedicated 15×15 SOM was built and trained so as to complete the three-cluster-based ride comfort assessment. This SOM is a particularisation of the one developed and analysed in the previous section, using only the variables of the input universe highly correlated with the fuel-consumption information, which, according to Table 1 are RMS(XACC pos) and RMS(ERPM).

In the same fashion as for the joint ride-comfort and eco-driving SOM, the maps of the weights, shown in Figure 10, can be used to assess and analyse how the SOM algorithm has extracted meaningful characteristics regarding fuel consumption.

Thus, it can be seen that both maps are different from each other, which means that the input variables have a low level of correlation between them. Regarding the weights displayed for RMS(XACC), according with Figure 10, uniformly decrease from left to right, with the maximum values laying on the upper and bottom left corners of the map. Concerning RMS(ERPM), it can be seen that the weights decrease from the bottom to the top, being this decrease more noticeable at the centre part than at the sides. In this case, the maximum values lay on the bottom-left corner of the map.

With the aforementioned characteristics of the input planes, we can assert in advance with no further analysis that the driving samples with the highest fuel consumption are those corresponding to the bottom-left quarter of the SOM. The medium and low consumption drivers, conversely, can be matched with the upper-left quarter and the right half of the map, respectively. However, since these latter trends are not as clear as the first one, a further analysis was performed by partitioning the SOM into 3 clusters, in the same manner as in the precedent section, as shown in Figure 11.

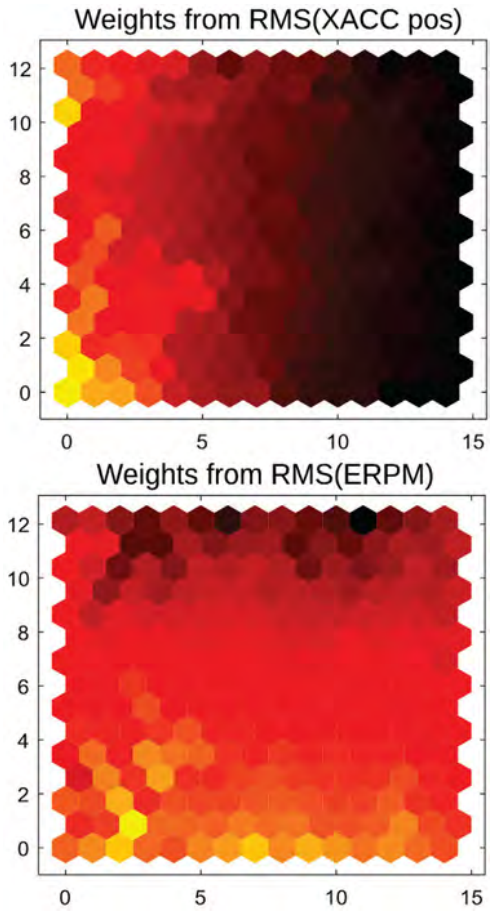

**FIGURE 10** Input weight maps for the two selected input features of the auxiliary SOM, RMS{XACC pos,ERPM}. A total of 1717 data windows were analysed. Lighter colours represent higher values of the corresponding features for each of the 15 × 15 neurons of the fuel-consumption SOM

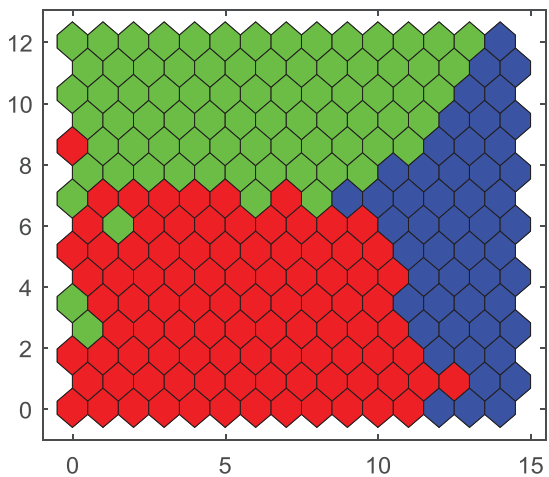

**FIGURE 11** Auxiliary SOM partitioned to separate neurons between three clusters of fuel consumption. This partition is also produced by carefully selecting the $k$ of the k-NN partitioning algorithm

**TABLE 4** Average values and variances for fuel consumption for the three-cluster auxiliary classification. This classification is performed on the total of 2290 windows

| Fuel consumption [$l$/100 km] | Blue | Green | Red |
| --- | --- | --- | --- |
| Avg. | 2.30 | 2.91 | 4.35 |
| Var. | 0.94 | 1.74 | 2.97 |

Three areas can be clearly distinguished in the shown clustering. These zones meet with the characteristics visually extracted through the previous paragraphs by the analysis of Figure 10 and consolidate the statements elaborated in advance. As depicted, the red cluster, which comprehends the bottom-left quarter of the SOM, joins drivers with high levels of RMS(XACC pos) and RMS(ERPM), which, according with the previously performed correlation analysis, match with the highest levels of fuel consumption. Regarding the green and blue clusters, they match with the previous assertions in terms of location within the SOM input planes. To verify that each class corresponds with individual fuel consumption categories, Table 4 was elaborated.

As compiled in the preceding table, the three clusters identified in Figure 11 clearly separate driving samples according to their fuel-consumption scores. Thus, the red cluster contains the samples with the highest consumption rating, being the green one representative of medium consumption, and the blue, those samples of the lowest consumption class. Hence, these results verify that if an auxiliary clustering of the fuel-consumption related variables is performed, a better separation of this feature can be achieved.

## 4.2 | SOM aggregated results

As previously stated, the elaboration of a main SOM to jointly assess ride comfort and fuel consumption individual characteristics did not provide an adequate level of eco-driving assessment. For that reason, a fuel-consumption exclusive, auxiliary SOM was developed. This dedicated SOM, in contrast with that for joint assessment, achieves a clear classification because only inputs with significant correlation with fuel consumption are applied. Thus, the joint evaluation of both SOMs is expected to enable to provide a robust classification of both characteristics.

Due to this approach, the operations of the driver on the commands of the vehicle are to be analysed for the ride comfort clusters first, and subsequently, for the fuel consumption groups. Finally, the operations for each objective will be joined to provide personalised recommendations to improve the DS-associated ride comfort and fuel consumption.

Thus, in accordance with Figures 10 and 11, it can be concluded that the different ratings happen because of noticeable differences on handling. The dataset provides two gas pedal signal, PGP and GP, both of which are highly correlated with the analysed feature XACC pos, being the correlation coefficient between XACC pos and PGP of 0.90. In the same way, the brake

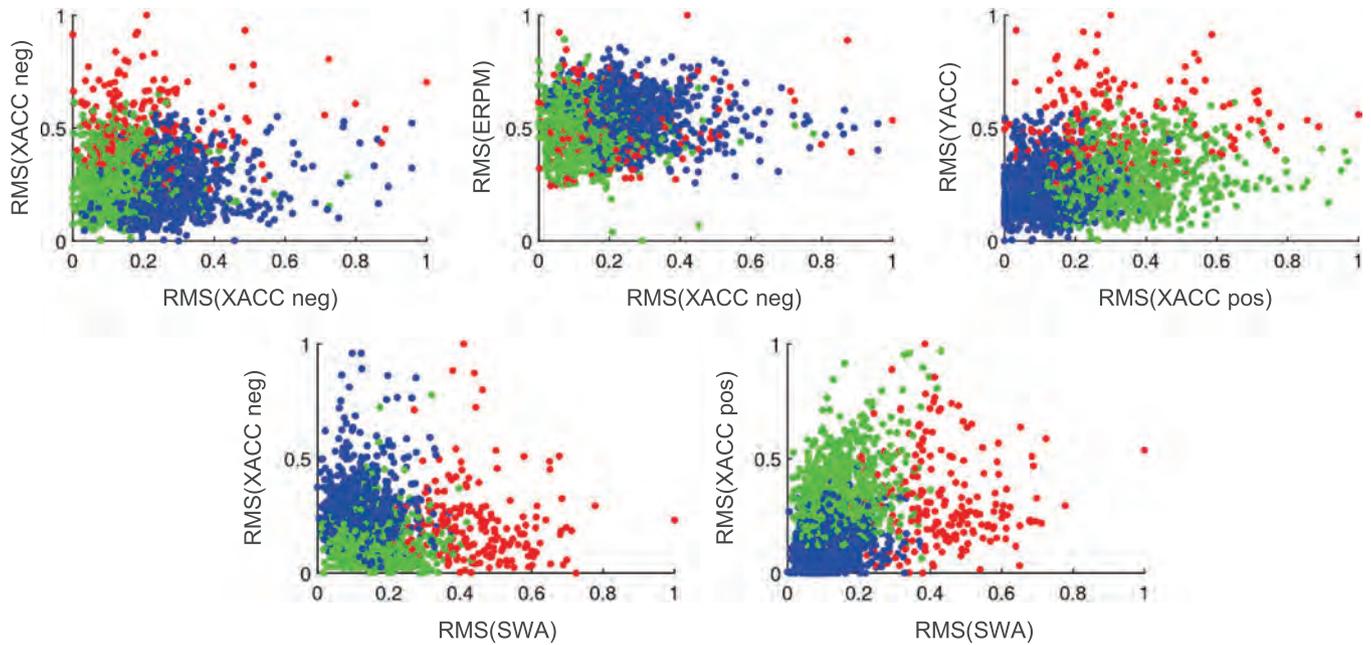

**FIGURE 12** 2D views of the three-cluster distribution for ride comfort. The clusters were labelled as high discomfort (red), medium discomfort (green) and low discomfort (blue)

pedal positively correlates with XACC neg, being 0.68 the correlation between them. So, it can be concluded that XACC pos and XACC neg are suitable features to carry out a qualitative assessment of the manner each driver operates the gas pedal and brake pedal, respectively. Finally, SWA and ERPM are representative features of the drivers' operation of the steering wheel and the gear stick, respectively.

## 5 | DEPLOYMENT OF THE DRIVER ADVICE MODULE

With the SOM-based in-car driving style modules properly developed and trained, achieving meaningful separation of the input samples into clusters, the eco-driving ride comfort improvement module is to be deployed, according to Figure 3. For this purpose, the causes of the ride comfort and fuel consumption characteristics of the selected clusters are assessed and meaningful recommendations are provided according to them.

### 5.1 | Ride comfort clusters' characteristics

According to Table 2 in Section 4, three ride comfort clusters are enough to distinguish drivers regarding their vomit rate and $MSDV_y$, being the red one the cluster with the highest rate, while the blue one represents the lowest. Intermediate motion sickness data is compiled into the green cluster. Thus, with the clustering corresponding with Table 2, Figure 12 is elaborated, and, with the depicted characteristics, we can classify the clusters as follows:

- High discomfort (red) corresponds to drivers with high values of vomit rate and $MSDV_y$, and, consequently, with elevated SWA and YACC, which means that the main feature of these motorists is that they tend to use the steering wheel aggressively, following swift cornering and overtaking strategies.
- Medium discomfort (green) drivers show moderate to low values of vomit rate and $MSDV_y$, and, consequently, with moderate to low SWA and YACC, while their values of XACC pos are medium-high, pointing out that the extensive use of the gas pedal is the main condition of ride comfort.
- Low discomfort (blue) class shows the lowest values of vomit rate and $MSDV_y$, so SWA is also kept in the lower range, XACC pos is kept low too, which means that these drivers tend to operate the gas pedal smoothly. Conversely, the values of XACC neg are higher than for the green class, which, jointly with the high correlation with nx neg, suggest that these drivers could use the brake pedal more thoroughly with spurious braking peaks.

Additionally, by the inspection of Figure 12, it can be seen that no separation is achieved for ERPM, which is coherent with the input planes displayed in Figure 8, and with the indetermination of the variance of the fuel consumption for each cluster.

Thus, according to the enumerated characteristics, as well as with the comfort variables displayed in Table 2 taken into consideration, the advice shown in Table 5 could be provided to drivers to improve their DS regarding ride comfort.

With this recommendations, and with the foregoing considerations in mind, we can elaborate Table 6 to show up the potential decrease on the likelihood of occupants to get motion sick

**TABLE 5** Suggested actions to improve ride-comfort

| Comfort cluster | Driver advice |
| --- | --- |
| High (red) | Operate steering wheel more smoothly |
| Medium (green) | Release gas pedal |
| Low (blue) | Avoid braking peaks[a] |

a)Note: This advice is only provided when a braking peak above a certain threshold occurs.

**TABLE 6** Expected vomit rate reduction between clusters

| Current cluster | Target cluster | Vomit rate reduction [%] | $MSDV_y$ reduction [%] |
| --- | --- | --- | --- |
| Medium (green) | Low (blue) | 9.88 | 6.62 |
| High (red) | Medium (green) | 46.0 | 54.7 |
| High (red) | Low (blue) | 51.3 | 57.7 |

in case of the DS recommendations were completely followed by the drivers.

As depicted in Table 6, a reduction of up to the 57.7% can be expected if a highly discomfortable driver could modify his/her DS so as to mimic a low motion-sickness motorist. Nevertheless, if he/she could only make it to drive like a medium motion-sickness driver, a promising reduction of up to the 54.7% could be achieved.

## 5.2 | Fuel consumption clusters' characteristics

As pointed out in Section 4.1, an auxiliary SOM was needed to alleviate the inconsistencies on the estimation of the fuel consumption for each of the ride comfort clusters. Thus, according to Table 4, three eco-driving clusters can be distinguished regarding the associated fuel consumption, with the red group representing the data with the highest fuel consumption, the green one, the intermediate consumption, and the blue, the lowest. With these clusters, Figure 13 was elaborated to assess the behavioural characteristics that cause the differences on eco-driving.

By the visual inspection of Figure 13, the three different clusters can be classified regarding their characteristics as follows:

- High fuel consumption (red) corresponds with samples with medium to high XACC pos and ERPM, with high dispersion, which means that these drivers tend to use the gas pedal swiftly and extensively at high engine regimes, with a low gear selected [71].
- Medium fuel consumption (green) drivers keep ERPMs too low, but the high average level of XACC pos indicates that the usage of the gas pedal is excessive while the gear selected is high.
- Low fuel consumption (blue) group keeps XACC pos in the lowest range while the ERPMs are kept within an adequate range [71].

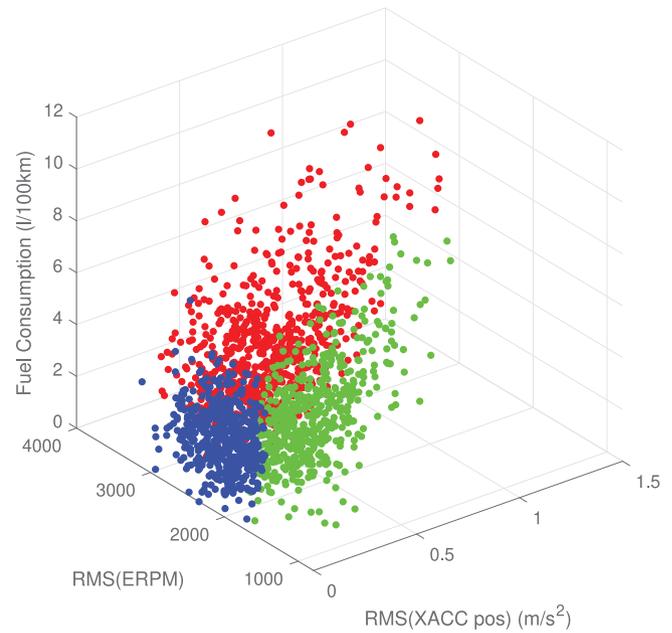

**FIGURE 13** 3D view of the three-cluster fuel consumption classification results. The clusters were labelled as high (red), medium (green), and low (blue), considering RMS(ERPM), RMS(XACC pos), and fuel consumption

**TABLE 7** Suggested actions to improve fuel consumption

| Comfort cluster | Advice |
| --- | --- |
| High (red) | Keep gas pedal steady / switch to a higher gear |
| Medium (green) | Release gas pedal / switch to a lower gear |
| Low (blue) | Keep driving style |

These automatically inferred groups reflect typical characteristics of several types of driving according to the "golden rules" of driving [58], where low RPMs and soft operation are required to achieve a low fuel consumption rate, while swift operation of the gas pedal jointly with high engine regimes is discouraged. According to the enumerated characteristics, the advice compiled in Table 7 could be provided to drivers to improve their DS regarding fuel consumption.

As depicted in Table 8, a reduction of up to the 47.1% can be expected if a high consumption driver could modify his/her DS so as to mimic a fuel-efficient motorist. Nevertheless if he/she could only make it to drive like a medium consumption driver a promising reduction of up to the 33.1% could be achieved.

**TABLE 8** Fuel consumption reduction between clusters

| Current cluster | Target cluster | Fuel consumption reduction [%] |
| --- | --- | --- |
| Medium (green) | Low (blue) | 21.0 |
| High (red) | Medium (green) | 33.1 |
| High (red) | Low (blue) | 47.1 |

**TABLE 9** Percentage distribution of the ride comfort and fuel consumption clustering intersection

|  | Low fuel consumption (blue) | Medium fuel consumption (green) | High fuel consumption (red) |
|---|---|---|---|
| Low discomfort (blue) | 24.3 | 1.51 | 1.86 |
| Medium discomfort (green) | 3.38 | 22.9 | 5.24 |
| High discomfort (red) | 15.9 | 20.1 | 4.83 |

## 5.3 | Joint recommendations

The recommendations provided in Section 4.2 suggest drivers to change the manner they operate some controls of the automobile, such as the steering wheel, the brake pedal, the gear stick or the gas pedal to mimic the behaviour of other motorists with better scores in either ride comfort or fuel economy. With the recommendations for these two separated clusterings taken into consideration, the intersection of the ride comfort and eco-driving groups if explored with the aim of providing the driver with a joint set of recommendations.

In Table 9 both clusterings are intersected. Columns represent the fuel consumption clusters, and rows, the ride comfort groups. As can be seen, the low discomfort group is mainly conformed by low consumption samples. The same happens for the medium discomfort cluster, shaped by medium fuel consumption drivers. Conversely, for the high discomfort class, two predominant fuel consumption classes, low and medium, can be identified.

Thus, Table 9 shows that a high correlation between comfort and fuel economy can be expected for the low and medium motion sickness classes, where most of the hits correspond with the same eco-driving classes. On the other hand, this correlation is not so clear for the high motion sickness class. This lack of definition of fuel consumption for the least comfortable class observed in Table 9 explains the high variance of the fuel consumption information observed for the joint clustering of Table 2.

With these considerations in mind, Table 10 was elaborated. It is worth to remark that, for the best driving style (i.e. low motion sickness and low fuel consumption) the driver is encouraged to keep on driving as he/she was, and advice is provided only when braking peaks occur. On the other hand, the least comfortable and eco-unfriendly drivers are provided with more advice.

Let us remark, as well, that the intersection of recommendations does not fall into contradictions, which proves the robustness of the system. Additionally, this lack of contradictions, as well as the fact of the recommendations being provided by means of natural language, makes the system more likely to be easily accepted by drivers, improving the engagement with the system, and, consequently, its effectiveness.

## 6 | SYSTEM ROBUSTNESS VERIFICATION

In this work it is shown that, with an appropriate selection of features, unsupervised machine learning methods can be used to extract conclusions according to the ride comfort and fuel economy standards regarding each driver's characteristics. For that purpose after the window size was properly sized and the driving signals were carefully selected regarding their Pearson correlation coefficients, two SOMs, one for the assessment of the ride comfort and another one for the fuel consumption were properly sized and trained. Finally, the classes identified by both SOMs were thoroughly examined to check that they did match with the already existing knowledge, and, analysing their members' values, recommendations according to their characteristics were elaborated and cross compared to create nine comfort-consumption clusters.

To prove that the intersection of both ride-comfort and fuel-consumption clusterings is robust and tends to classify drivers according to the kernel density estimation displayed in Figure 7, heat maps for drivers 2, 6, 8, 14, 18, and 20 were elaborated. Figure 14 represents the distribution of the cross classification for each of the drivers. If these distributions are compared with their corresponding kernel density estimation surfaces (Figure 7), it can be seen that drivers with the sharpest kernel density distribution, such as drivers 2 and 6, tend to show a predominant medium–medium discomfort+fuel consumption cluster, driving during more than the half of the samples in this class. In contrast, for drivers such as drivers 8, 14, or 18

**TABLE 10** Intersection of the ride comfort and fuel consumption recommendations

|  | Low consumption (blue) | Medium consumption (green) | High consumption (red) |
|---|---|---|---|
| Low discomfort (blue) | Keep driving style<br>Avoid braking peaks[a] | Release GP / Lower gear<br>Avoid braking peaks[a] | Keep GP steady / Higher gear<br>Avoid braking peaks[a] |
| Medium discomfort (green) | Keep driving style<br>Release GP | Release GP / Lower gear<br>Release GP | Keep GP steady / Higher gear<br>Release GP |
| High discomfort (red) | Keep driving style<br>Operate SW smoothly | Release GP / Lower gear<br>Operate SW smoothly | Keep GP steady / Higher gear<br>Operate SW smoothly |

[a]Note: This advice is only provided when a braking peak above a certain threshold occurs.

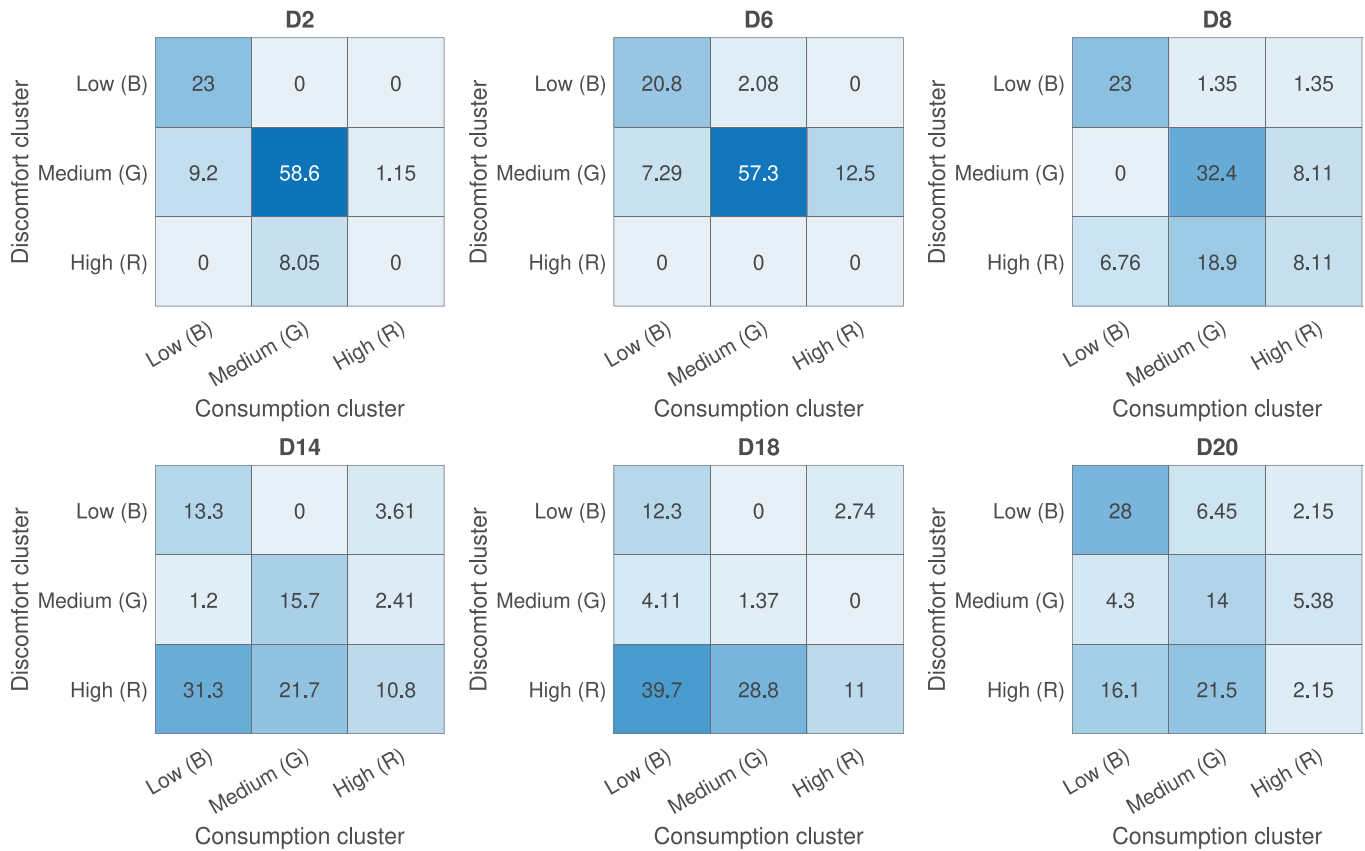

**FIGURE 14** Heatmaps of the percentage distribution of the intersection of the ride comfort and fuel consumption clusters for drivers 2, 6, 8, 14, 18, and 20. B stands for blue, G for green, and R for red

the distribution is more disperse, with several clusters having a high predominance, lacking a single predominant driving cluster.

On the other hand, conclusions about either ride comfort (i.e. by rows) or fuel consumption (i.e. by columns) can be extracted by analysing Figure 14. Thus, the predominant cluster for driver 2 is medium, with the 68.9% of his/her windows laying into this category. Regarding fuel consumption, it can be seen that medium is also the winning classification, with 66.7% of his/her windows being classified for this category. Driver 6 behaves similarly to driver 2, the most frequent combination is also medium-medium, with 77.1% and 59.4%, respectively. For driver 8, medium-medium, with 40.5% and 52.7%, is again the winner combination. But, in this case, the highly discomfortable class as well as the low consumption class are also remarkable, with the 33.8% and the 29.7%, respectively. Drivers 14 and 18 are very similar, with the highly discomfortable class outstanding among the others, with the 63.9% and the 79.5% of the driving windows, while most of the times they are classified as low consumption drivers (45.8% and 56.2%). Finally, driver 20 shows a fairly uniform distribution, so, this driver cannot be clearly classified separately for fuel consumption or ride comfort. In contrast, the class intersection shows nine clearly separated categories that can be easily distinguished, proving that the followed approach is useful to perform a deeper insight into driving behaviour, specially for drivers with no predominant class.

Regarding the recommendations displayed in Table 10, they should be provided in a noticeably enough way, but without being confusing or annoying for drivers, helping to achieve an engagement level with the system such that the expected improvement ratios could be effectively reached. For that reason, specially with the aim of minimising confusion, recommendations are only to be provided when DS is clearly and steadily identified during a certain time threshold, avoiding the annoyance of displaying messages that constantly change, and, consequently, improving the effectiveness and friendliness of the system.

## 7 | CONCLUSION

The main motivation of this work was the development of an ADAS to increase ride comfort of the passengers, while taking into account the eco-driving viewpoint. The proposed solution provides the driver with a set of recommendations, in a natural language, with the aim of improving his/her driving style. The system is composed of two main subsystems: a SOM-based driving style classifier and the driver advice module. The first one, consists of two cooperative SOMs that were trained with real-world driving data. The classifier subsystem is able to group drivers regarding their ride comfort and eco-driving characteristics. The second subsystem identifies the underlying causes of both the DS-associated lack of comfort and high consumption, and provides advice according to them. The verification carried out on the whole system proved that it is robust enough to clearly identify the aspects of ride comfort and eco-driving jointly in most real-world driving situations, providing a deeper insight than if each aspect were analysed separately.

The aforementioned recommendations are designed to be easily understandable by most of the drivers, and, if they were completely followed, noticeable reductions in the comfort compromising parameters could happen. As shown in this work, if a driver could modify his/her DS with the help of the recommendations from the most discomfortable group to the most comfortable one, the discomfort indicators would improve up to the 57.7%, improving the comfort perception of the vehicle occupants drastically. As for eco-driving, if a motorist achieves a very high level of engagement with the system, modifying his/her DS according to the advice provided, drastic reductions of up to the 47.1% in both fuel consumption and GHG emissions could be achieved.

### ACKNOWLEDGEMENTS
This work was supported in part by the Spanish AEI and European FEDER funds under Grant TEC2016-77618-R (AEI/FEDER, UE), by the University of the Basque Country under Grant GIU18/122 and by the Basque Government under Grant KK-2019-00035-AUTOLIB (ELKARTEK).

### CONFLICT OF INTEREST
The authors declare no conflict of interest.

### DATA AVAILABILITY STATEMENT
Restrictions apply to the availability of these data, which were used under permission.

### ORCID
*Ó. Mata-Carballeira* https://orcid.org/0000-0002-1468-6280
*I. del Campo* https://orcid.org/0000-0002-6378-5357